# Complementing the Linear-Programming Learning Experience with the Design and Use of Computerized Games: The Formula 1 Championship Game

**Gerardo L. Febres** [1]

[1] Departamento de Procesos y Sistemas, Universidad Simón Bolívar, Sartenejas, Baruta, Miranda 1080, Venezuela; gerardofebres@usb.ve

**Abstract:** This document focuses on modeling complex system situations to achieve an advantage within a competitive context. Our goal is to devise computerized games' characteristics to teach and exercise non-easily quantifiable tasks crucial to the math-modeling process. We split the math-modeling process into three layers and present the games' essential characteristics to fulfill these purposes. We also introduce a computerized game to exercise the math-modeling process and optimization problem formulation. The game is named The Formula 1 Championship, and models of the game were developed in the computerized simulation platform MoNet. It resembles some situations in which team managers must make crucial decisions to enhance their racing cars up to the feasible, most advantageous conditions. This paper describes the game's rules, limitations, and five Formula 1 circuit simulators used for the championship development. We present several formulations of this situation in the form of optimization problems. Administering the budget to reach the best car adjustment to a set of circuits to win the respective races can be an approach. Focusing on the best distribution of each Grand Prix's budget and then deciding how to use the assigned money to improve the car is also the right approach. In general, there may be a degree of conflict among these approaches because they are different aspects of the same multi-scale optimization problem. Therefore, we evaluate the impact of assigning the highest priority to an element, or another, when formulating the optimization problem. Studying the effectiveness of solving such optimization problems turns out to be an exciting way of evaluating the advantages of focusing on one scale or another. Another thread of this research directs to the meaning of the game in the teaching-learning process. We believe applying the Formula 1 Game is an effective way to discover opportunities in a complex-system situation and formulate them to finally extract and concrete the related benefit to the context described.

---

## 1. Introduction

Courses of optimization, specifically linear programming, are now a classical field forming essential content of several career study plans. During the last four decades, linear programming techniques have been the core subject of operations research and linear model courses in undergraduate and graduate studies. The typical study plan for these courses includes formulating linear optimization problems and acquiring techniques to solve these problems with computers' help.

The use of games to study the outcomes of relationships in a complex competing environment was pioneered by H. Stackelberg, who published his book "Marktform und Gleichgewicht" in 1934, translated to English in 2011 with the title "Market Structure and Equilibrium" [1]. Stalkelberg's model became the so-called Stalkelberg game and was one of the Nested Optimization's first formalizations; today better known as Bi-level optimization. The use of games to create situations where the players get experience and feel the consequences of their decisions was fostered with the advent of digital computers and, specifically, with the spreading of computerized games. Among the first in this category was the Beer Distribution Game [2]. As part of the material supporting the Systems Dynamics, Jay Forester, now considered the father of this field. The Beer Game demonstrates how quickly a few interacting systems parts raise the complexity of the system's dynamics. The Beer Game has been a successful in-class activity still used today all over the world to teach systems dynamics and, in some cases, as the basis to promote changes in the teaching methods [3].

The use of games as an educational tool has intensified during the last decades. The Outcome-Based-Education, introduced by Spady in 1994 [4], proposes changing the education system's paradigms by moving the center of the learning process from the teacher to the students. The student-centered learning process can be strengthened through computerized games. In 2011 Tracy Sitzmann presented a study [5] summarizing "the literature on the instructional effectiveness of computer-based simulation games for teaching work-related knowledge and skills." In her work, Sitzmann concludes the self-efficacy of trainees working with simulation games was 20% higher than a reference group non-working with simulators. Sitzmann's conclusion is better understood, mentioning the meaning of self-efficiency: "confidence in the proper capacity to achieve goals or to learn effectively" (B. Zimmermann [6], and B. Zimmermann, A. Kitsantas and A. Campillo [7]). In their papers, H. Wideman et al. [8] and M. Coovert et al. [8] highlight the appropriateness of games as a research tool and better prepare students for their future role as professionals.

Linear optimization serves as a platform to illustrate applications and the usefulness of the broader subject of optimization, filling it with valuable content more comfortable to follow for abstract careers as engineering, management, and urban sciences. While these practices led to the vital reward of spanning linear programing techniques all over the managerial sciences, it seems the non-less important aspect of detecting opportunities for applying these techniques has always been aside.

In administering resources, the application of games as a teaching tool has not been exploited. Classical courses of Operations Research focus on algorithms and techniques but leave out of the scope to recognize opportunities. We do not know of a study program that systematically provides formation on the ability to create ideas about the use of limited resources so that these limitations become opportunities within the context of a competitive environment.

Our goal is to develop ways to stimulate the ability to discover opportunities in the real complex and competitive world. Thus, this paper aims to present a game based on a five-race Formula 1 Championship simulation.

The organization of this paper is into four parts. The first explains our view of the math-model process and games' characteristics to exercise the modeling process effectively. The second contains a description of the simulators used for the game of the F1 Championship. The third part analyzes different approaches to focus on optimizable situations leading to the formulation of optimization problems. The last segment is devoted to present results and a brief discussion.

## 2. Teaching more than techniques.

As a method to convey lessons with content beyond only techniques, competition games are the main proposal of this paper. Competition games seem to be particularly suited for subjects with particular emphasis on mathematical modeling. While mathematical modeling is unavoidably part of any engineering activity, there are some specialties where the mathematical model represents most of the product sought. Engineering Management, Operations Research are just two examples of such engineering branches. Thus there are many subjects in the engineering career which products result being mathematical models, and paradoxically, out of the math-technique component, the quality of a mathematical model proves to be hard to quantify. Evaluating a math-model by only considering its syntax, which is part of what we call 'technique,' would be like assessing an English text by considering only its grammar and orthography. Originally math-models, or math, exist because of our need to measure and quantify. Later, math became a language we also use to record and transmit ideas in the most precise way we can. Math is more than numbers, counting, arithmetic, algebra, equations, spaces, and patterns. Math is a universal language serving us to grow in our understanding. Ironically, math itself is unable to measure the overall quality of a math-model. There will always be elements in our models lying outside the possibilities for precise evaluation.

As a method to convey lessons with content beyond only techniques, competition games are the main proposal of this paper. Competition games seem to be particularly suited for subjects with particular emphasis on mathematical modeling. While mathematical modeling is typically only a part of any engineering activity, there are some specialties where the mathematical model represents most of the product sought. Engineering Management, Operations Research are good examples. Moreover, study plans of subjects with an intense application of math-modeling conventionally include the most necessary techniques. However, they do not include mechanisms to teach how to achieve successful results in real-life scenarios. No objective assessment of the quality of the model is pursued. We see three reasons for explaining these issues.

- The difficulty of quantifying the actual quality of a mathematical model and forecasting its effectivity in a real future situation.
- The problems typically used in optimization courses follow a characteristic formulation pattern. These problems do not cause the students to struggle for a good, creative, and competitive result.
- The problems' answer is considered right if it complies with correctly applying formulation techniques and model-writing syntaxes. Providing students with rigid steps toward the formulation is the typical method used to teach these techniques.

During the last decades, teaching has accentuated the importance of Outcome-Based learning [4]. Outcome-Based learning provides a way to overcome the difficulties of grading complex answers to complex situations the students have to "solve" when demonstrating learned capabilities. Furthermore, the availability of powerful computers led to the use of games as a tool or vehicle to implement a virtual —but real— situation to immerse the students into a real competence context. T. Sitzmann [5], B. Zimmermann [6], B. Zimmermann, A. Kitsantas and, A. Campillo [7], H. Wideman et al. [9], and M. Coovert et al. [8] presented results in this regard.

More recently, this discussion's focus is moving toward the model as one of the names of the structures we rely on to think. Therefore, models appear now as essential 'objects' when we strive to improve our thinking methods. The research by Katehi, Linda, Greg Pearson, and Michael Feder (2009) [10] shows that early stimulation of math and modeling skills at pre-school stages considerably increases the "Habit of Mind in Engineering" and fosters the students' future understanding of engineering systems. In 2017 Lammi M. and D. Cameron D.[11] present an interesting approach about the role of modeling in the "Habit of Mind in Engineering." Lammi and Cameron set up study cases to examine the engineering process toward the design of some artifacts. Their results highlight the fact design activities do not follow a linear procedure. We regard these observations as applicable even when the engineering product is a math model instead of an artifact.

*2.1. The math-modeling process*

Our conception of math modeling goes beyond the strict scope of syntactical and technical math-language considerations. Engineers in real-life situations usually require the modeling of complex systems. This activity is a compound of several steps that occur in a recursively and non-independent way.

We see three distinctive layers forming the overall math modeling process.

- Understanding problem objective: Identifying the essential problem parameters. This phase typically requires management and business skills to identify the key aspects leading to a measurable advantage.

- Select the approach and tools to apply. In this layer, the analyst sets priorities and the general model structure. Here the analyst establishes the scale —in the sense of Febres G.L. [12]— to focus the model. Experience and heuristics are crucial in this phase. Experience and heuristics are crucial to this phase.
- Describe the model with math-language and numerically solving it. This step represents the most technical modeling layer and is the one always included in linear optimization courses. This layer comprises mathematically formulating the problem, feeding it into a computer to find the numerical solution and analyze the results.
- In the more general case of modeling and design, a fourth layer complements the former: physical prototype construction and commissioning.

With this math-model description in mind, our goal is to define games' characteristics to make the students strive to complete the tasks involved in the three layers of the math-model processing.

*2.2. Game requirements*

A game should comprehend the following features to fulfill the diverse activities involved in our teaching plan:

1. Several scales of behavior. Scale in the sense of Febres [12]. The optimization game should contain several parameters to be adjusted to realize an overall advantage. These parameters may 'exist' at a specific scale within the situations represented in the game. Conveniently, optimization parameters exist at different scale levels so that the students have to select, compare, or weigh the convenience of focusing around one or another scale of the game.
2. Considerable complexity. Although any multi-scale will most likely be sufficiently complex, we must emphasize complexity is an essential condition to create games that resemble real-life situations.
3. A precise measure of the overall system-model performance and an optimizable situation. Despite the modeled system's complexity, there must be a recognizable, unarguably, and quantifiable performance index. It must be a unique index.
4. Approximately linear behavior in the most relevant aspects. This feature is required only to be consistent with the essence of linear-programming courses. For other mathematical modeling courses, we estimate this condition is not required.

A competitive environment offers conditions for knowledge extent and exercising originality and creativity. It also provides a natural way to rank the quality of the models presented by each student group participating in the course.

**3. The Formula 1 Championship Game**

Implementing the Formula 1 Championship Game aims to make the students face up the need for improving a complex system's performance towards an optimized use of the resources. The Formula 1 Game is a complex system whose performance depends on the interaction of several subsystems. Figure 1 shows a schematic of the Formula 1 Game's main components seen as a system with its major components.

The game consists of a championship comprising five Formula 1 Grand Prix. The participating teams must use an available budget while seeking the car's enhancement to finish the contest successfully. As in the real championship, each race gives points to the first ten positions at the end of the race. After the five races, each team's number of points indicates the final positions of the contest.

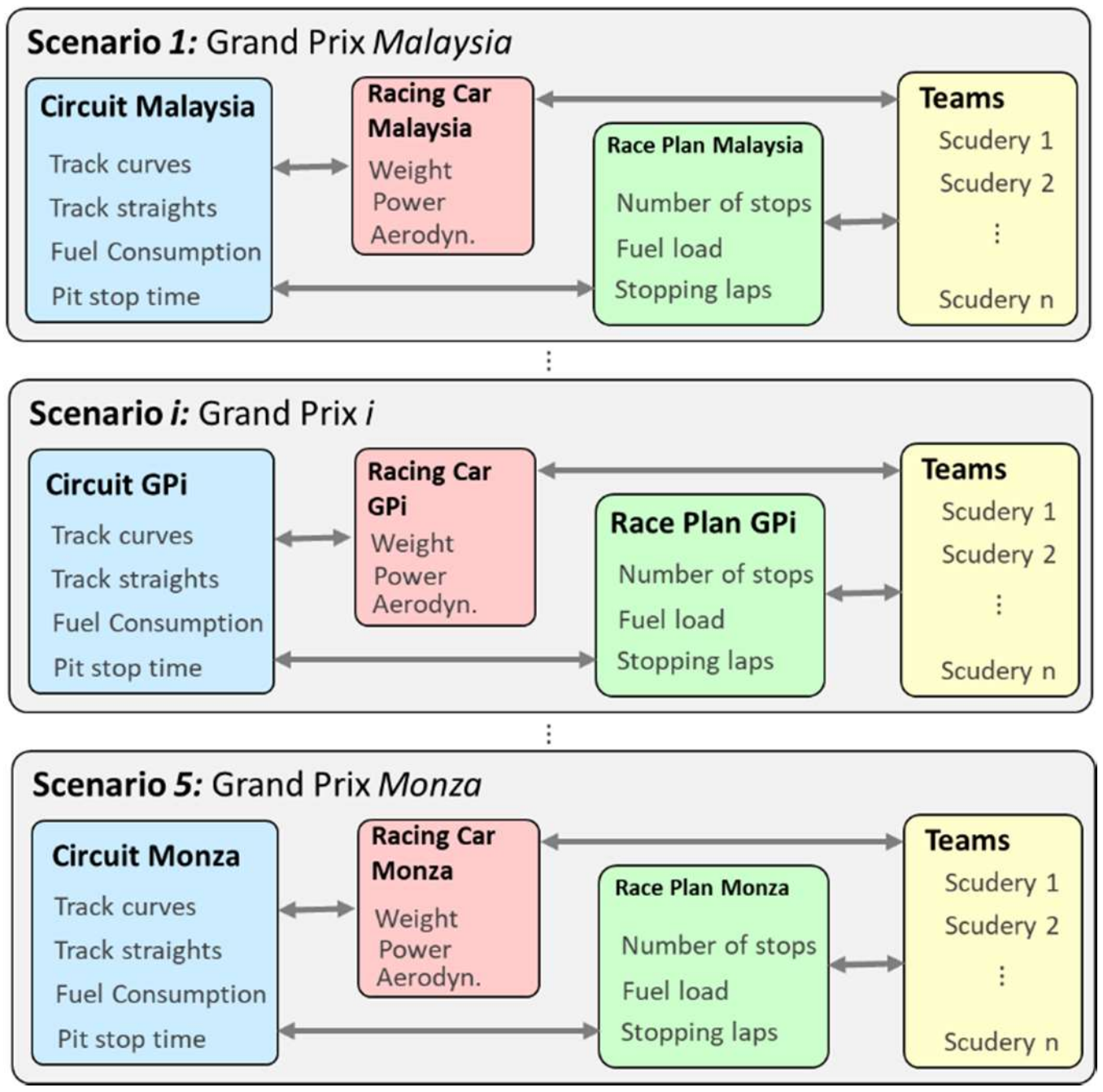

**Figure 1.** Schematics of the Game. Each Grand Prix is modeled as a system comprising the interactions of several subsystems. Arrows represent direct interactions. Thus two subsystems not connected with an arrow are modeled as mutually independent.

We depict some aspects of the game model and the associated simulators in the next sections.

### 3.1. *Car perfomance model*

A mathematical model represents the behavior of the racing cars. The racing cars model consists of three performance parameters: the horsepower, the weight, and the aerodynamic-suspension parameter. These parameters can be improved at the expense of a certain amount of money that cannot exceed a budget. A set of starting parameters describe the initial car for all teams. We have chosen the following values: aerodynamic-suspension parameter = *1*, horsepower = 830 HP, and weight = 702 Kg. The car weight was chosen low to accentuate the impact of the fuel's consumption and the associated car's weight reduction as the race progresses. We add a set of constraining conditions $Ax$, $Px$, and $Wx$ corresponding to the aerodynamic-suspension parameter, the horsepower, and the weight to enhance the car is a bounden process.

### 3.2. *Circuit model.*

Each circuit model is a geometrical model. The model comprises a sequence of curved or straight components. The depiction of straight elements consists of its distance, while the curves require the radius and the rotation angle. All circuit parts share the same track width, which is also a circuit model parameter. The circuits are modeled as flat; there are no banked curves and no height differences between any two circuit sectors. Figure 2 shows a comparison of a satellite view Malaysia circuit with a schema of its representation within the model.

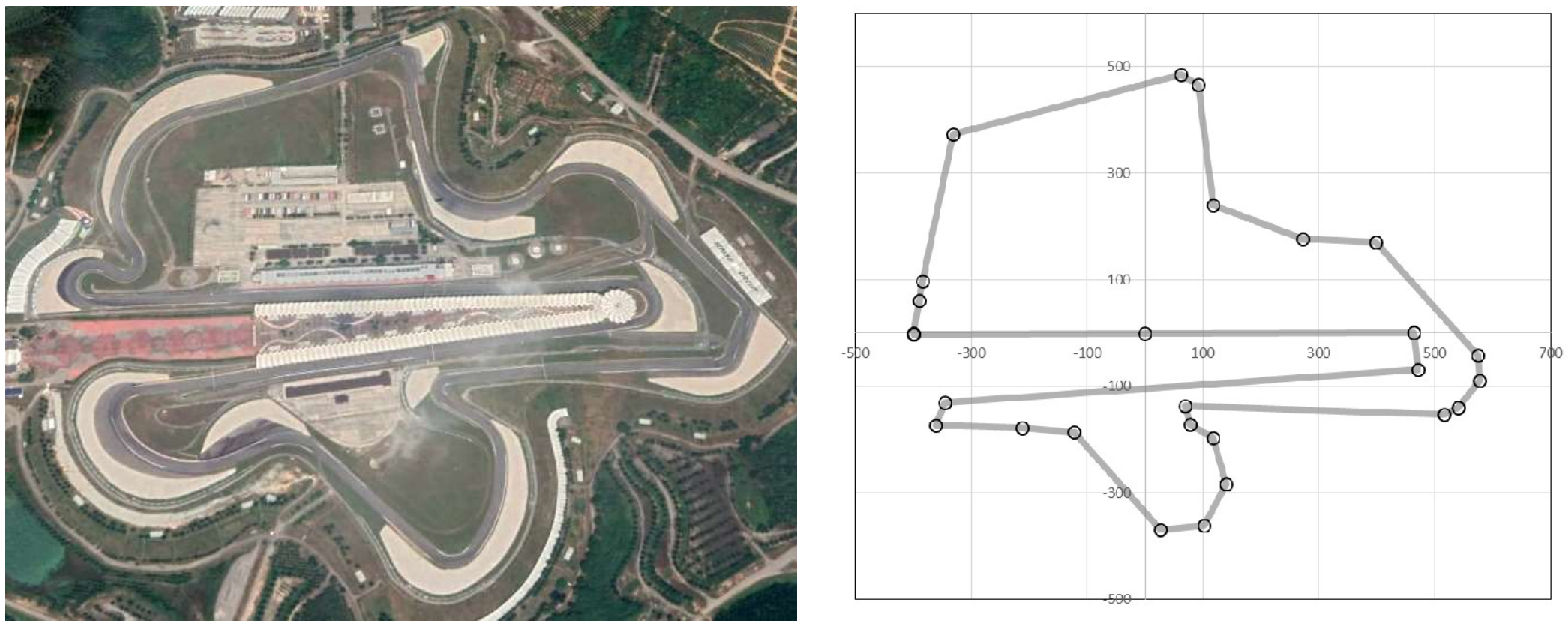

**Figure 2.** The Malaysia circuit. On the left, a satellite view of the circuit (taken from Google Earth on 2020.01.30). On the right, the Malaysia circuit representation obtained from the simulation model. The bubbles show the points where circuit elements, straights or curves, connect. The curves appear as straight lines; however, in the actual model, they are segments of circumferences with the radius adjusted to connect the neighbor elements tangentially.

### 3.3. *Trajectory model.*

An exact model to represent the best feasible trajectory for a curve is a difficult task. The selected trajectory depends on factors like driving style, the curve bank, the effectivity of the brakes (which we are not considering here), the car weight distribution that may change as the fuel is consumed along with the race. Thus, defining the best trajectory for a curve is a very subtle matter that depends on a detailed analysis of each specific situation. For this study, we consider the fastest trajectory in a curve is the one that minimizes the centripetal force required to keep the car within track. Thus, independently of the speed, car mass, and friction of the tires with the track, the fastest trajectory must be close to the maximum possible radius trajectory. Figure 3 shows a schematic drawing of a generic curve.

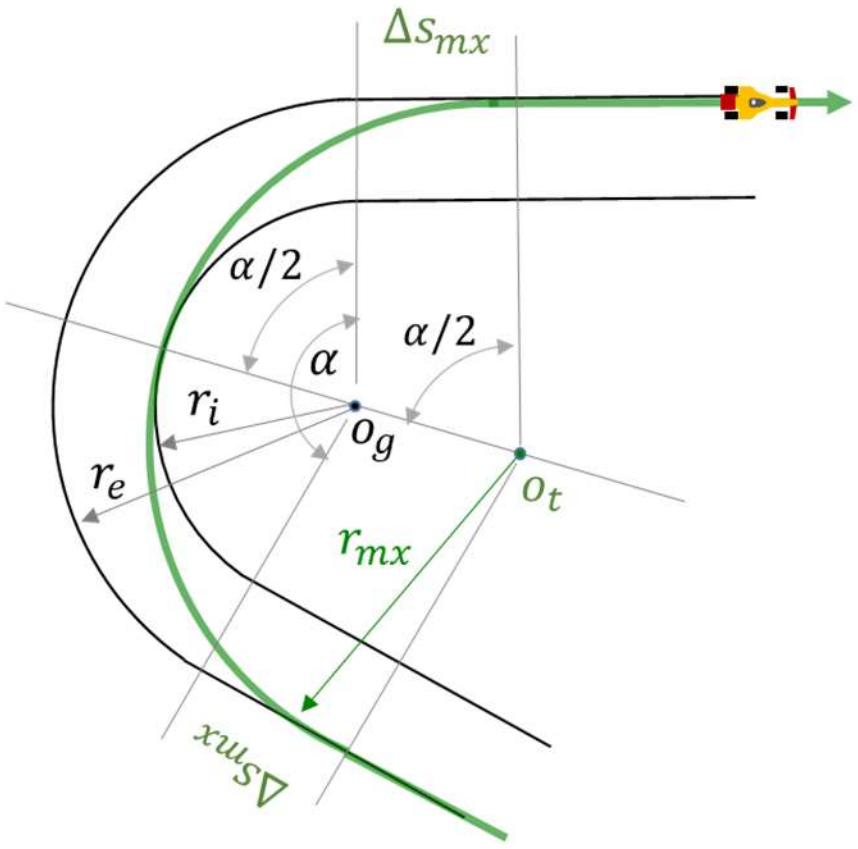

**Figure 3.** Geometric treatment of a curve with internal radius $r_i$ and external radius $r_e$. The curve's rotation angle is $\alpha$. The point $o_g$ represents the geometrical center of the curve and the point $o_t$ represents the center of the maximum radius trajectory, which is shown with the green line. The term $\Delta S_{mx}$ indicates the length of the straights surrounding the curve converted into part of the curve and thus taken away from the effective straight length.

The schematic on Figure 3 supports the formulation of Equation (1) to obtain an expression for the fastest radius of a curve.

$$r_{mx} = \frac{r_e - r_i \cos {}^{\alpha}/_{2}}{1 - \cos {}^{\alpha}/_{2}} \quad for\ 0 < \alpha < \pi\,, \tag{1}$$

$$r_{mx} = r_e \qquad\qquad for\ \pi \le \alpha\,.$$

Likewise, the length of straight segments absorbed by the curves is computed with expression (2).

$$\Delta s_{mx} = (r_{mx} - r_i) \sin {}^{\alpha}/_{2} \quad for\ 0 < \alpha < \pi\,, \tag{2}$$

$$\Delta s_{mx} = r_{mx} - r_i \qquad for\ \pi \le \alpha\,.$$

*3.4. The simulations.*

The simulations consist of computing the time circuit-lap for the car model's conjunction and its performance improved conditions. The simulation allows indicating the weight of fuel loaded before the completion of a set of laps is run. In the actual Formula 1 races, there is no refueling in the middle of a race. However, in this game, we split the race into two or three sets of laps to let each team refuel the car. Another difference with the real case is that we do not include the change of tires in these simulations.

A lap's time is the addition of a car's times to run all straight segments and all curves contained in the circuit. It is assumed the race car goes thru a curve at the limit speed allowed by the aerodynamic conditions combined with the curve's geometrical conditions. Reference maximum speed for all curves are set following information taken from circuit schematics and information available on the internet through Keith Collantine's web site RaceFans [13].

The straight segments are modeled assuming a constantly accelerated movement starting with the previous curve's speed until the race car must stop accelerating because of the next curve or until it reaches the maximum circuit speed indicated by specialized and updated information sources as [13]. For the accelerating segment, we use the model presented in expression (3).

$$V_f - V_s = C_p\, L\, \frac{H}{W}\,, \tag{3}$$

where $V_s$ and $V_f$ are the start and final speeds at the straight segment, $L$ is the effective length, $H$ and $W$ are the horsepower and the weight of the racing car, and $C_p$ is a coefficient adjusted with reference values to reproduce a reasonably good fit to the impact of the horsepower, the weight, and the length for all straights of the circuit over the speed gain after running the straight segment. The reason for modeling the straight segment by using these coefficients is that we prefer building a linear model. The best lap time and other performance measures can be determined by applying linear programming techniques.

*3.5. Modality of the races*

Each simulated Grand Prix race consists of the same number of laps to the circuit as its real counterpart. The competing cars all start from the start-finish line. Thus, there is not such a thing as the starting grid.

In the Grand Prix model, up to two pit stops are allowed for refueling. The additional time that a pit-stop-lap appears in the simulator files. Even though refueling is not part of the current real Formula 1 competition, including it in our contest adds complexity to the problem, thus better resembling real-life situations. Deciding when to stop for refueling makes up an additional level of optimization. Conversely, our Formula 1 game does not include changing tires as part of the system studied in the present form. Adding the requirement of using more than one set of tires, as the current real championship rules indicate, would increase the difficulty of selecting a better direction in which an opportunity will be found. Requiring several sets of tires is one of the most attractive changes to be implemented in futures versions of the game.

*3.6. The budget*

Each team has a budget to be used along with the championship. The money is to pay for the improvements of the race-car performance parameters. The costs for the upgrades of a performance parameter is proportional to the magnitude of the improvement. However, these costs may vary upon the position of the race within the championship chronogram. Table 1 shows the races' order and the costs of a 1% improvement of each performance parameter applied during the last championship, which occurred in November of 2019.

**Table 1.** Costs of improvements of 1% for each car performance parameter. These are the values applied during the Championship of 2019's last trimester

| Cost of enhancing the car's performance parameters [$] | | | | | | |
|---|---|---|---|---|---|---|
| **Parameter** | **Symbol** | **1st GP** | **2nd GP** | **3rd GP** | **4th GP** | **5th GP** |
| Aerodyn-Suspension | $Ba_i$ | 700000 | 600000 | 500000 | 450000 | 400000 |
| Horsepower | $Bh_i$ | 120000 | 100000 | 90000 | 85000 | 80000 |
| Weight | $Bw_i$ | 150000 | 120000 | 100000 | 90000 | 80000 |

*3.7. General restrictions*

A variety of restrictions define the boundary of the feasible space for a prospective optimization problem formulation. In the current game, the maximum budget is 6 million dollars. The number of pit-stops can be up to two. Refueling pit-stops are not compulsory, but the fuel tank capacity is limited to 70 Kg. The car's attribute enhancements are also limited. Table 2 shows a synthesis of the conditions restricting the current situation. By changing the limits of restrictions, we can set many different game versions. The difficulties may range from a very easy to formulate problem to a multilevel optimization problem.

**Table 2.** Limiting conditions for the Formula 1 Championship game, version 2019.

| Limiting conditions | | |
|---|---|---|
| **Condition Name** | **Symbol** | **Value** |
| Budget | $B$ | 6 million dollars |
| Racing car max. Aerodyn-suspension parameter | $Amx$ | 1.09 |
| Racing car maximum horsepower | $Hmx$ | 1025 HP |
| Racing car minimum weight | $Wmx$ | 650 Kg |
| Minimum number of refueling pit-stops | | 0 |
| Maximum number of refueling pit-stops | | 2 |
| Fuel tank maximum allowable load | | 70 Kg |
| Racing car minimum weight | | 650 Kg |

*3.8. The game*

Before each Grand Prix, the teams must submit to the championship stewards (the professor) the car performance adjustments, the costs of these expenses, as well as the weight of gasoline put before the start of each race sting and the lap number of each refueling pit stop. The steward then uses the simulators to compute each team's race time and publishes all groups about the Grand Prix results. The published results include the times of the race stings and the pit-stop lap number used for refueling. The judge does not include how each team distributed the money among the performance parameters and the weight of gas put before the stings in the published results.

The simulators used in the Formula 1 Championship do not have noise or randomness in any of the expressions used. Thus every car-circuit simulator is a deterministic model. However, it is impossible to predict the precise development of the championship because all teams can, taking into account the results of the previous races, adjust their bid for the upcoming races. All players know others' goals, but their exact strategy is uncertain. Thus, this championship fits within the definition of a game.

After every Grand Prix, points are distributed as follows: 25 points for the first place, 17 points for the second place, 12 points for the third place, nine points for the fourth place, seven points for the fifth place, two points for the eighth place, and one point for the ninth place.

### *3.8. Implementations*

The game relies on a set of five simulated Formula 1 competitions. Using a spreadsheet environment, we built a set of simulators for the circuits of Malaysia, Silverstone (England), Las Americas (USA), Monza (Italy), and Sochi (Russia). These spreadsheet-based simulators are available in [14]. For about four years, spreadsheet versions of the Formula 1 Game allowed reaching the game's concept's maturity. They successfully tested its application with the participation of engineering and urban science students. The Formula 1 Game can be generalized for other educational games, research, and even other entrepreneurial applications. However, using the Excel version is very time-consuming since it requires feeding the data and running the game for each combination Team-Grand Prix. Simulating a system as the conjunction of several subsystems and using these digital structures as a sort of "routines" that receive racing-cars, race-plans, and circuit components as "arguments" is just beyond any software's capabilities we currently know. Thus there is the need for a tailored simulation computerized platform.

The recently developed simulation platform Monet [15] complies with the requirements of the modeling environment for the Formula 1 Game. Monet is a modeling platform with strong capabilities representing multidimensional entities and to model systems at several observation scales. Currently, we implemented models for the five circuits of the championship in Monet. This game's new version [16] includes the contracting of the driver as an additional variable parameter

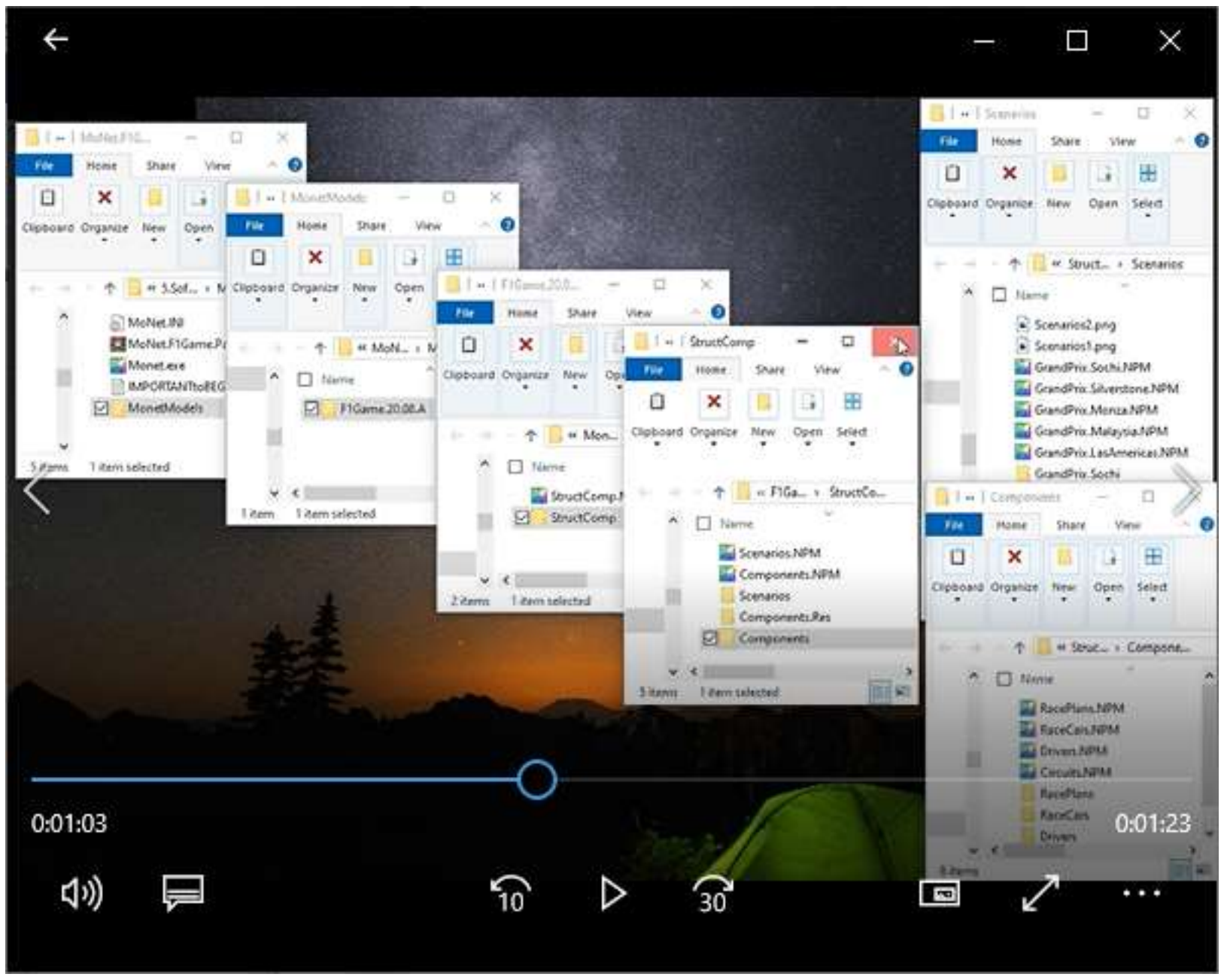

**Figure 4.** The F1Game file-structure. Detailed exposition in video available in [17].

The use of MoNet-based Formula 1 game is a system of component models. There is a model for each type of component of the game system. There are also several instances for each type of component model. Thus, for example, a racing car is described by its weight, horsepower, and aerodynamic condition for each Grand Prix race and each competitor team. Similarly, a race plan indicates when the lap a pits stop occurs and how much refueling corresponds for each race for each participating team. Combining and keeping track of these "complex variables" at each stage of the championship is a rather complex task requiring a tailored developed system as MoNet and the computer's correspondingly complex file-structure. Figure 4 presents a screen capture with several directories showing subdirectories and files integrating the whole Formula1 championship game. A video showing the size and extent of this file-structure is available at [17].

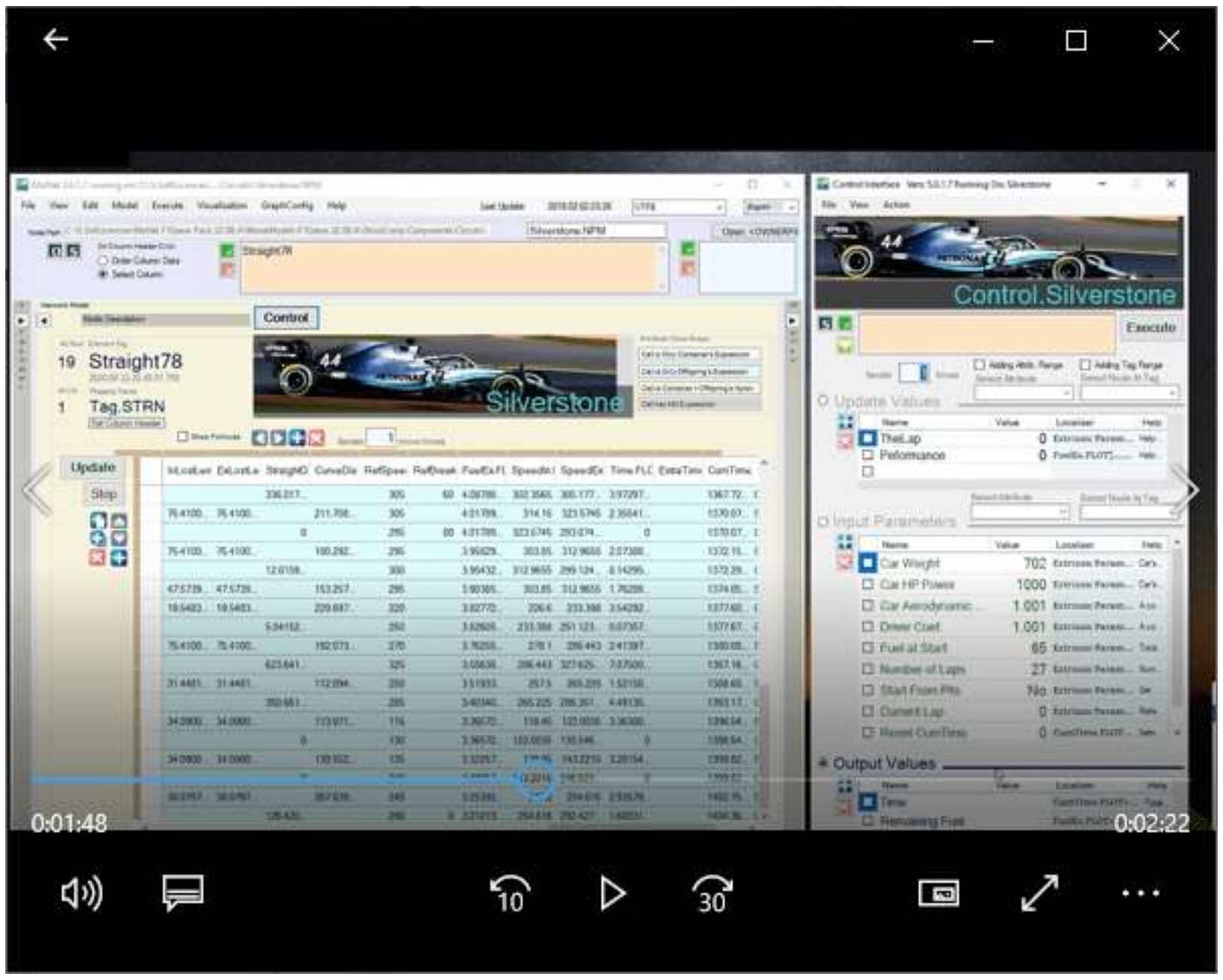

**Figure 5.** Testing racing cars in a MoNet-simulated circuit. Detailed exposition in video available in [18].

In the Formula 1 Championship Game, the use of MoNet occurs from two perspectives. The student-competitor perspective mainly uses the circuit models to make the necessary tests to gather data, letting them build a specific strategy to win the contest. Figure 5 presents a window containing Monet's description of a circuit and a window dedicated to controlling parameter values to be tested by the contesters. A video showing the use of MoNet to make these tests is available at[18].

The other mode of using the MoNet-based game is the teacher -facilitator-judge perspective. This use mode is spreader than the competitor's use but does not require making any decision, just registering and administrating results. The facilitator's tasks include feeding the competitor's related data into the system's components and asking the system to compute the results for each segment of each team and race of the championship. Figure 6 shows a screenshot of the video, including some tasks to account for a team's time in a race.

Supporting files, including Videos 1, 2 and 3, are available in [17–19]. The new version of the game will be launch in the next optimization course period starting in September 2020. Due to the current pandemic conditions, it is appropriate to deploy new online activities in the upcoming engineering courses.

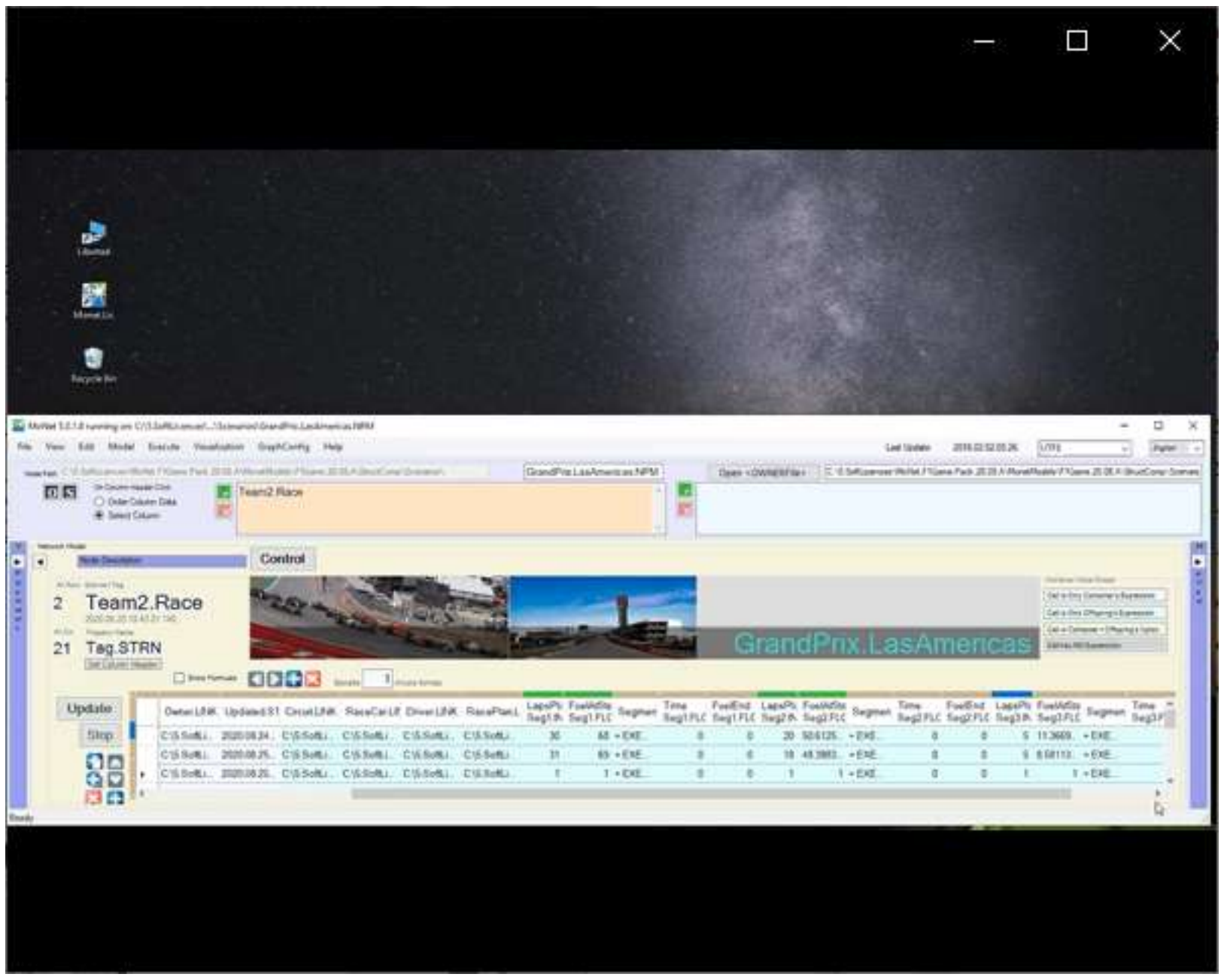

**Figure 6.** Accounting results in the F1 MoNet-based game. Detailed exposition in video available in [19].

## 4. Optimization models

The Formula 1 game is a simplified model of the situation race directors and team-managers face in real life. Most model components have a close-to-linear behavior, and their number is substantially reduced concerning the real case. Nevertheless, the presented model is realistic because a change introduced in a parameter produces a reaction in the expected direction of the directly related parts of the systems. However, there is not an obvious or easy way to distribute the budget into the five races and the three car-performance parameters to obtain a performance improvement profile leading to winning the championship. Moreover, since any decision about the expenses devoted to the car-performance parameters comes after deciding the amount given to each Grand Prix, this situation may be a Bi-level optimization problem. Thus, implementing optimization models is a valuable method to use the budget in the best possible way.

### *4.1. Formulation at the level of races. RLF.*

An optimization model the students frequently identify to reach a good competitive condition is allocating resources to enhance the racing car performance parameters: horsepower, weight, and suspension-aerodynamic parameter. Adjusting the car to each circuit, expending the most effective distribution budget, is an obvious path to increasing competitiveness. Problem (3) presents a formulation of this model for the race identified with the sub-index $i$. The idea is implemented by changing the budget dedicated to improving the suspension-aerodynamic coefficient, the horsepower, and the weight, which the variables $xa_i$, $xh_i$, and $xw_i$ represent. The coefficients $Sa_i$ , $Sh_i$ , and $Sw_i$ describe the improvement of a lap-time gained by expending $xa_i$, $xh_i$, and $xw_i$ on the suspension-aerodynamic factor, horsepower, and weight. Therefore, the objective function returns the total time reduction obtained with the budget expenditures. The parameters $Amx$ , $Hmx$ , and $Wmx$ constrain the suspension-aerodynamic coefficient, the horsepower, and the weight, respectively. The values $Av_i$, $Hv_i$, and $Wv_i$ represent the suspension-aerodynamic coefficient, the horsepower, and the weight after the previous race and before the improvements implemented for the current competition. The total expenditure devoted to improving the car's race $i$ performance cannot exceed the amount $B_i$ dedicated to that race.

$$max \quad Sa_i \cdot xa_i + Sh_i \cdot xh_i + Sw_i \cdot xw_i \,, \tag{4}$$

$$\text{subject to:} \quad Sa_i \cdot xa_i \leq Amx - Av_i \,,$$
$$Sh_i \cdot xh_i \leq Hmx - Hv_i \,,$$
$$Sw_i \cdot xw_i \geq Wmx - Wv_i \,,$$
$$xa_i + xh_i + xw_i \leq B_i \,,$$
$$xa_i \,,\ xh_i \ ,\ xw_i \geq 0$$

The bound values $Amx$, $Pmx$, and $Wmx$, are directly given as part of the racing car description and are easily interpreted. The values $Sa_i$, $Sh_i$, and $Sw_i$ are, in turn, elusive to some students, perhaps due to the fact they must develop a little experiment with each circuit simulator to extract these values.

*4.2. Formulation at the level of the championship. CLF.*

Another perspective to make decisions about the use of the budget is to expend the money accordingly to the impact an expenditure has o the different circuits, to wit, spending the budget where the money is more effective. Problem (5) shows a formulation directed to maximize the overall impact achieved by expending the budget $B$ along the $M$ races comprising the championship.

$$max \quad \sum_{i=1}^{M} \alpha_i \cdot I_i \cdot B_i \ , \tag{5}$$

$$\text{subject to:} \quad \sum_{i=1}^{M} I_i \cdot \leq B \,,$$
$$B_i \geq 0 \ \text{for } i = 1 \text{ to } M \,.$$

The impact $I_i$ of each dollar expended in race $i$ can be established using the simulators and then distributing the amount $B_i$ assigned to race $i$ in predefined fractions to each car's performance parameter. Problem (5), as formulated, includes in its objective function the factor $\alpha_i$, which considers that the budget early spent in the championship has effects for the car's better performance in more races than the budget spent late in the championship. The determination of an appropriate distribution of factors $\alpha_i$ is not an obvious task and easily becomes a matter of discussion. The result is that, while most teams detect the possibility of applying the optimization approach at the level of the championship, very seldom they come up with a clear idea about how to formulate it. Frequently, $\alpha_i$ values are set by applying diverse reasonings that are far from the actual set of optimal values.

*4.3. Optimizing refueling pit-stops.*

In this Championship version, up to two refueling pit-stops are allowed. Every time a car stops for refueling, an additional partition appends to the race. The transit by the pit lane implies a time that is added to the corresponding partition. On the other hand, avoiding a pit-stop means running the car with an extra fuel weight that affects the circuit laps' time. Comparing the pit lane additional time versus the additional time caused by the extra fuel weight needed to avoid the pit stop determines the best number of pit-stops. Most importantly, this aspect of the problem can be decoupled from formulating the rest of the optimization problem.

*4.4. A comprenhensive multi-level formulation. CMLF.*

A model incorporating the previously explained approaches into a single and complete actual optimization model fits into bi-level optimization problems. Recently Ankur Sinha, Pekka Malo, and Kalyanmoy Deb [20] offered a review of bi-level optimization problems. Rigorous formulation of bi-level problems is not within the course scope where we use this Formula 1 Championship game. However, exercising the ability to recognize opportunities for squeezing the available resources in complex and challenging model processes is an essential course goal. None of the student teams has presented a comprehensive formulation to play the Formula 1 Championship game in our experience.

Nevertheless, we suggest a model for this problem that assesses the goodness of the student teams' role during the course experiences. We start by depicting the objective function by saying it must represent the 'time gained' using the available budget to enhance the car performance parameters. Then, we must define a suitable time reference to compute the "time gained" later. The race time that the original (unchanged) car would take to complete each race is the set of references we select. Therefore, the time gained is for Grand Prix $\boldsymbol{i}$ is the juxtaposition of the improvements achieved on three car performance parameters. For race $\boldsymbol{i}$, the time gain $\boldsymbol{g_i}$ depends on the sensitivity of expenditure of money on enhancing a performance parameter and the car's improved condition after being improved for the previous races. Computing the time gain $\boldsymbol{g_i}$ by scaling the expenditures the varying costs of the improvement of a 1% of performance parameter for previous races yields the following expression:

$$g_i = \sum_{k=1}^{i} \left( \frac{Ba_i}{Ba_k} \cdot Sa_i \cdot xa_k + \frac{Bh_i}{Bh_k} \cdot Sh_i \cdot xh_k + \frac{Wh_i}{Wh_k} \cdot Sw_i \cdot xw_k \right) \quad , \tag{6}$$

which we use as the base for the objective function. To complete the formulation, we need to avoid the time gains concentration in a small subset of races. The Formula 1 Championship is the typical situation justifying the use of goal programming techniques. A group of arbitrarily selected weights establishes the priorities of the time gained in each race. We prefer to apply an alternative method that considers the expected tendency of the time gains all teams are prone to obtain during the Championship development. The growth pattern of time gains is likely to be exponential due to the multiplicative superposition of each race's performance enhancements. A general model for the time gain of a race $\boldsymbol{g_i}$ in terms of a reference number $\Delta \boldsymbol{G_{ref}}$ and the base $\lambda$ of an exponential number is then:

$$G_i = \Delta G_{ref} \cdot \lambda^{i-1} \quad , \tag{7}$$

Expression (7) is a useful tool since it allows representing a naturally exponential process into a rigorously linear optimization model. To select the value of $\lambda$, the average of the enhancement factors for the races is a good guide for the choice of $\lambda$ values. For this study we used $\lambda = \mathbf{1.15}$. The value of $\Delta \boldsymbol{G_{ref}}$ must be selected to maintain a reasonably large feasible space, so the optimization model has enough freedom to select a competitive strategy. Finally, Problem (8) presents the Comprehensive Formulation.

$$max \quad \sum_{i=1}^{M} \sum_{k=1}^{i} \left( \frac{Ba_i}{Ba_k} \cdot Sa_i \cdot xa_k + \frac{Bh_i}{Bh_k} \cdot Sh_i \cdot xh_k + \frac{Wh_i}{Wh_k} \cdot Sw_i \cdot xw_k \right) \tag{8}$$

| | |
|---|---|
| $subject\ to: \quad \sum_{i=1}^{M} xa_i + xh_i + xw_i$ | Budget |
| $Av_1 = 1, Hv_1 = 830, Wv_1 = 702,$ | First race condition |
| $Av_i + Ca_i \cdot xa_i \leq Amx \quad , for\ i = 1\ to\ M,$ | Aerodyn. enhancement |
| $Hv_i + Ch_i \cdot xa_i \leq Hmx \quad , for\ i = 1\ to\ M,$ | Power enhancement |
| $Wv_i + Cw_i \cdot xa_i \leq Wmx \quad , for\ i = 1\ to\ M,$ | Weight enhancement |
| $\Delta G_{ref} = 0.5, \lambda = 1.15$ | First race enhancement |
| $Sa_i \cdot xa_i + Sh_i \cdot xh_i + Sw_i \cdot xw_i \geq \Delta G_{ref} \cdot \lambda^{i-1}, for\ i = 1\ to\ M,$ | Time gain pattern |
| $xa_i, xh_i, xw_i \geq 0 \quad , \quad for\ i = 1\ to\ M,$ | Non negativity |

Problem (8) is a simplified formulation presented here to avoid a bi-level problem's rigorous formulation. The objective function is the sum of the reductions of the lap-time achieved by enhancing the $M$ races' car performance parameters. These lap-time reductions are modeled by multiplying the amount of money dedicated to each performance parameter by the sensitivity ($S\ a_i$ , $S\ h_i$ $or$ $S\ w_i$) to each dollar expended on it. Set of constraints keeps the feasible space from trespassing the allowed limits of the performance parameter enhancements. An additional set of constraints serves to avoid the sum of time reduction concentrating on a few races. At any race $i$, the enhancement must not be lower than a number defined by $\Delta G_{ref}\ \lambda^{i-1}$.

*4.4. Heuristical and unstructured formulations. H- and U.*

Some teams do not follow any of the formulations described above. We use two terms to classify these contest-propositions: heuristical and non-structured. Thus, if the resource to be optimized is identified, but the methodology used is based on rules or believes, we call this approach a heuristic and add a letter "H" to the resource's level being considered. Therefore, if the championship budget is distributed by considering heuristical reasons, we classify this "formulation" HCLF. Similarly, suppose the "formulation" does not follow a formal structure. In that case, we refer to the strategy by adding the letter "U." We never recommend setting up these ineffective strategies. Nevertheless, the occurrence of some heuristic and non-structured pseudo-formulations provides interesting data to evaluate the effectiveness of applying proper and legitimate optimization formulations.

*4.5. Other optimization-levels for formulation*

During each race, several aspects may impact the resulting race time. In the last championship version, the amount of gasoline loaded at the start of the race and the laps selected for refueling represent decisions that must occur one before the teams submit their race plan. Before making these decisions, the teams should evaluate the gas consumption process, studying whether it is related to other conditions present in the race. Whatever the evaluation result is, this aspect of the problem constitutes a different, typically finer, optimization level.

**4. Results**

The first version of the game, released during the year 2016, required some car physical performance modeling from the students. Optimizing the distribution of performance enhancements was only applied in one Grand Prix. Progressively, the game evolved toward a multi-race championship, while the car-circuit simulation model was retired from the students' responsibility. In its current condition, the game is a guide designed to convey the learning that opportunities are present in almost any real-life situation and that conceiving a multi-level formulation is an excellent method to take advantage of those opportunities. For the last version of the Championship, the costs of the improvements appear in Table 1. Table 2 shows a budget of 6 million dollars and other limiting conditions.

In evaluating our experience incorporating this game into the course plan, we address the results in two necessary scopes. One is the contest's numerical results to get a sense of how effective the optimization technique is. The other is to estimate the students' achieved stimulus about the relevance of efforts directed to the conception of valuable optimization models and their mathematical formulation.

Table 3 and Figure 3 refer to the results of the last Championship. The relevant conditions for this study's effects, as race time and points won for each team, are included in Table 3. The number of pit stops is not presented, but we can mention that almost all groups stopped twice for races two to five, while for race one (Malaysia), the most convenient number of refueling pit-stops is one. According to the formulation types depicted above, the second left column of Table 3 shows the class to which the teams' strategy belongs. The first row of Table 3 shows the time that an original (not enhanced) racing-car would take to complete the run for all the Championship circuits' distance. We computed these times considering a typical fuel load and the number of pit-stops.

**Table 3**. Type of formulation applied, race completion time (in seconds), and points won for the Championship teams.

| | | Malaysia | | Silverstone | | Austin | | Monza | | Sochi | |
|---|---|---|---|---|---|---|---|---|---|---|---|
| Team | Form. | Time [s] | Pts. | Time [s] | Pts. | Time [s] | Pts. | Time [s] | Pts. | Time [s] | Pts. |
| Ref. | - | 5118.08 | | 5004.96 | 1 | 5347.65 | | 4621.07 | | 5880.66 | |
| 1 | URLF | 5063.76 | 5 | 4930.86 | 1 | 5151.87 | 2 | 4255.18 | 3 | 5560.95 | 12 |
| 2 | HRLF | 5043.35 | 7 | 4898.14 | 9 | 5063.64 | 17 | 4239.92 | 9 | 5590.19 | 2 |
| 3 | CLF | 5019.91 | 12 | 4948.43 | 0 | **5029.67** | 25 | 4225.37 | 17 | 5577.24 | 7 |
| 4 | HRLF | 5060.02 | 6 | 4916.06 | 3 | 5095.41 | 5 | 4232.71 | 12 | 5574.46 | 9 |
| 5 | U | DNF | - | 4911.87 | 7 | 5209.61 | 2 | 4315.36 | 1 | 5534.00 | 17 |
| 6 | CLF | DNF | - | **4872.92** | 25 | 5070.72 | 12 | **4213.97** | 25 | 5589.27 | 3 |
| 7 | RLF | 4981.32 | 17 | 4892.26 | 12 | 5093.84 | 7 | 4243.02 | 7 | 5611.57 | 1 |
| 8 | U | 5042.41 | 9 | 4925.82 | 3 | 5235.40 | 1 | 4273.23 | 2 | 5577.61 | 5 |
| 9 | U | DNF | - | 4930.17 | 2 | DNF | - | DNF | - | DNF | - |
| 10 | U | 5067.60 | 3 | 4913.99 | 5 | 5090.56 | 9 | 4243.38 | 5 | 5593.24 | 9 |
| 11 | CMLF | **4943.29** | 25 | 4877.11 | 17 | 5122.05 | 3 | 4296.73 | 2 | **5507.30** | 25 |

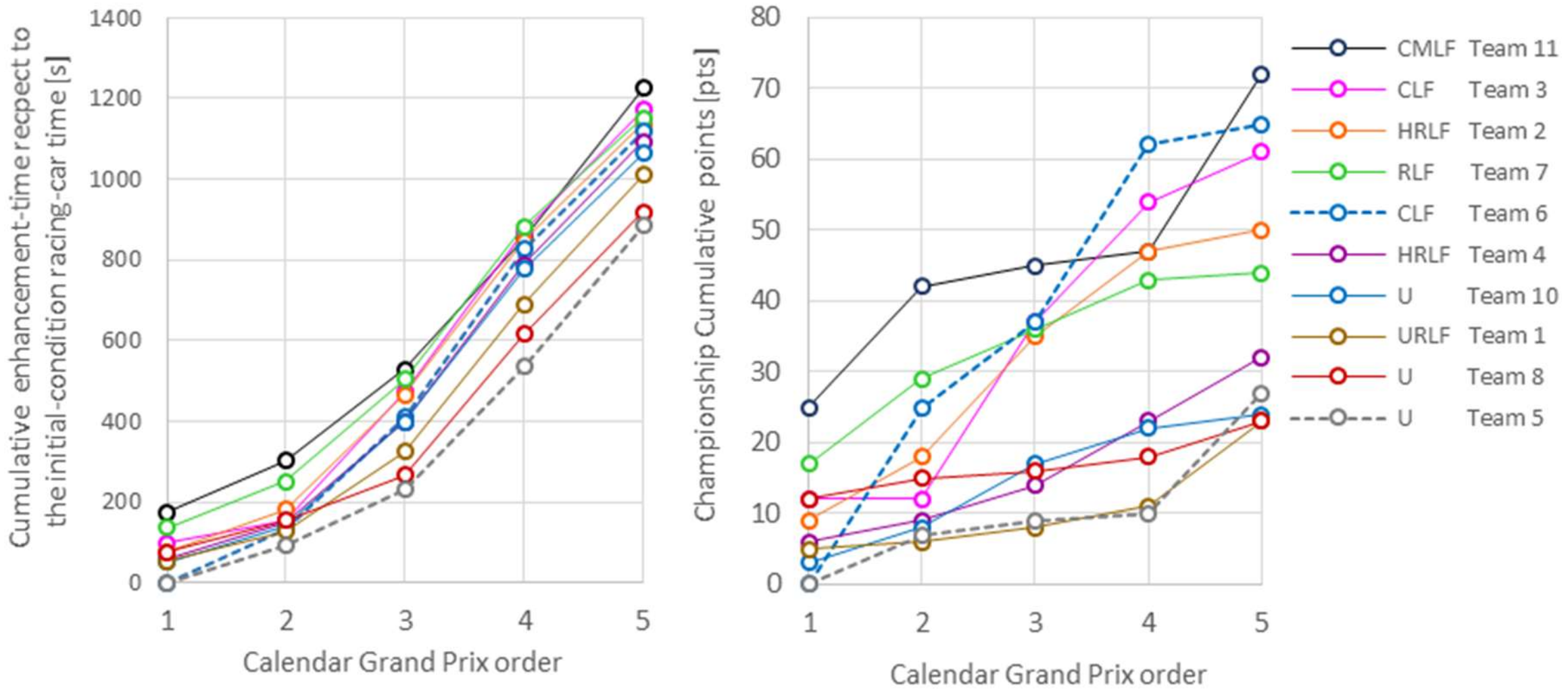

**Figure 7.** The Championship results shown in a chronological sequence of races represented in the horizontal axis. On the left graph, the vertical axis represents the time reduced respect to the time of the initial reference car. The time is computed as seconds accumulated over the progress of the races. Higher values in the vertical axis mean better performance. The right graph shows the cumulative points won by each team over the development of the races. Colored lines represent competing teams. Dotted lines correspond to teams not finishing all races. The legend also shows the optimization strategy followed by each team.

Team 11 was not present during the 2019's contest. It is added in this paper for comparative purposes representing the formulating model shown in Problem (8) and the formulation CMLF. Specific adjustments were made for the formulation of Team 11 until the values $\lambda = 1.15$ and $\Delta G_{ref} = 0.5$ settled down.

Figure 7 presents the results of the races of the Championship. To provide a sense of the teams' progression of race performances, we plot the cumulative differential times between the race-time and the reference original-car time. The higher the differential time is, the better the team's position in the race results. The representation of the cumulative differential times adjusts to the sense of objective functions formulated in problems (4), (5), and (8). However, the champion is determined by accounting for the number of points obtained during the Championship, which holds a highly un-linear relationship with the race time registered. Therefore, we add Figure 7 right showing the cumulative points for each team for the Championship.

**Table 4**. Estimated influence of the type of formulation.

| Impact of type of formulation | | | | |
|---|---|---|---|---|
| | Cumulative enhanced time | | Championship points | |
| **Type of formulation** | **Abs. [s]** | **Fraction** | **Abs. [pts]** | **Fraction** |
| CMLF: Comprehensive | 24746.48 | 1.00 | 72 | 1.00 |
| CLF: Championship level | 24800.62 | 0.96 | 61 | 0.85 |
| RLF: Race level formulation | 24822.01 | 0.94 | 44 | 0.61 |
| H: Heuristic | 24856.95 | 0.91 | 39.5 | 0.55 |
| U: Unstructured | 24975.415 | 0.81 | 25.25 | 0.35 |
| Reference | 25972.42 | 0.00 | 0 | 0.00 |

The graph lines associate the cumulative time's evolution with the progress of the Championship for each team. The optimization models used are classified as the closest of the models described in this document's former section. Figure 7-left exhibits a systematic pattern of cumulative times for all teams represented. Figure 7-right, on the contrary, shows the associated cumulative points that look disorganized and less predictable. This difference is perhaps due to the distinctive meanings of the vertical axes. In Figure 7-left, the vertical axis is the value of the objective functions of formulations RLF, CLF, and CMLF, thus producing soft and organized value patterns. The vertical axis of Figure 7-right represents a value connected to the objective function's value with a highly non-linear relationship.

Table 4 shows the overall results. The data collected may not be abundant, yet it seems sufficient to obtain an estimate of the achievable benefits of applying a well-conceived formulation criterion.

**5. Discussion**

We have been developing the Formula 1 Championship Game since the year 2016. Early versions of this activity required implementing some modeling of the physical aspects of the car-circuit system. Thus, the simulation was part of the course. This schema evolved toward a specialized activity centered on optimization techniques and the exercise of recognizing opportunities.

Before the races, the teams do not know what the 'next move' of the other teams will be. However, they know all other groups will do their best to consider the contest rationally' time-perspectives and win the Championship. Therefore, the Formula One Championship is formally a game, and thus forecasting an outcome is almost impossible. Simultaneously, selecting a good strategy or selecting the best approach does not guarantee to win the game.

Notwithstanding the complexity of this game, we think the data collected from plying this version of the game demonstrates the importance of formulating optimization problems compared to the use of heuristic methods. There is no doubt about the value of applying optimization algorithms to pursue better results in game-situations like the case now exposed. As expected, the Comprehensive Multilevel Optimization Formulation best exploits the resources to maximize the opportunities to win this game. On the other hand, unstructured approaches and formulations based on heuristics demonstrating not being able to complete with well-reasoned optimization formulations. Within the structured formulations, the ones considering the higher level of optimization seem to have an advantage over those focused on the lower-scaled simulated conditions.

The punctuation system applied in the game, selected for its similarity to the actual Formula 1's punctuation system, is highly un-linear. The accentuated number of points given to the first places of a race may have dramatic consequences over the team-order after the Championship's end. While all races cover similar distances and distribute the same number of points to the winners, the non-linearity mentioned makes it valuable to win a few races and disregard others. To wit, the distribution of points introduces a significant distortion on the proposed linear objective function. Under this circumstance, the game behaves like a non-convex problem. A capricious location of the objective function's surface is more likely to win than the place where any optimization algorithm, linear or not, would locate the optimal solution point. Several students warned about the highly non-linear punctuation system's effects during the last version of this game. They signaled the inconsistency of keeping this punctuation system in a game that a linear response in all aspects should dominate. We shall point out that this time the un-linear distribution of points accentuated the advantage of properly formulating the problem.

In 2007, B. A. Husein [21] classified the simulation games as Functional simulation games or Leadership simulations games. We believe the Formula 1 Game belongs to a different class of games: the Sharpness simulation games. This type of game aims to exercise creativity and to organize resources in a controlled and precise fashion. Thus, we regard the Formula 1 Championship Game as a vehicle for sharping the student's capacity to figure effective solutions in a complex, competitive scenario.

## 6. Conclusions

For four years, we have developed an optimization game. We use the Game as a complementary activity within our Linear-Optimization course at Simon Bolivar University, Venezuela. The Game started as a simulation activity but has evolved towards a structure that exemplifies situations where optimizing-based decisions are challenging, primarily due to the competitive context in which the Game presents. There are plans to extend the Game onto Decision Making and Optimization Problem Formulation for undergraduate and graduate studies. Other games within the same style but representing business and infrastructure development situations are in the plans to be created and applied in future courses.

The Formula 1 Championship Game is now a useful tool for selecting what aspects to focus on a developing phenomenon within complex systems and practice our mental models' organization. For the students, this game's outcome is the very experience of being exposed to similar real-life situations, where effort and systematic thinking lead to the effective use of the available tools, enabling them to take advantage of the opportunities they create.

The Formula 1 Game, apart from being a joyful challenge, depicts a scenario that, despite being complex, offers quantitative measures of the performance of a player. This situation persuades the student about the advantages of applying these techniques and justifies formal optimization strategies. Conveying knowledge should go beyond teaching technics. Easy to say but much harder to concrete. For centuries, the teaching activity has unquietly been developing ideas to extend our thinking's dimensionality and identify aspects that may be relevant while being shadowed for their minor scale. We think there is substantial progress within this perspective. We will remain looking for paths to effectively teaching thinking.

**Availability of data and materials:** The simulators developed in commercial spreadsheets and the current game components developed over the simulation platform MoNet, are available as indicated in the References Section.

**Funding:** No external founding was received for this research.

**Conflicts of Interest:** The author declares no conflicts of interest.

**Acknowledgments**: I wish to thank the students of the subject Linear Models (PS-1111) dictated in the Bolívar University during 2016, 2017, 2018, and 2019 for their enthusiastic participation in the game. I am pleased to give a special mention to Rommel Llanos, Cesar Rosario, Samuel Fagundez, Arturo Yepez, Luis Alvarado,

Mariela Castro, Guillermo Lopez, Josman Peralta, Andres Gonzalez, Amin Arriaga, Javier Medina, Christian Oliveros, Antonella Requena, Sandra Vera, Gianfranco Ravelli, Ivanova Marin, Jorge Aponte, and Koraima Galavis for their best effort in the contest and valuable ideas they provided for the adjustments of the current version of the Formula 1 Championship Game.

## Appendix

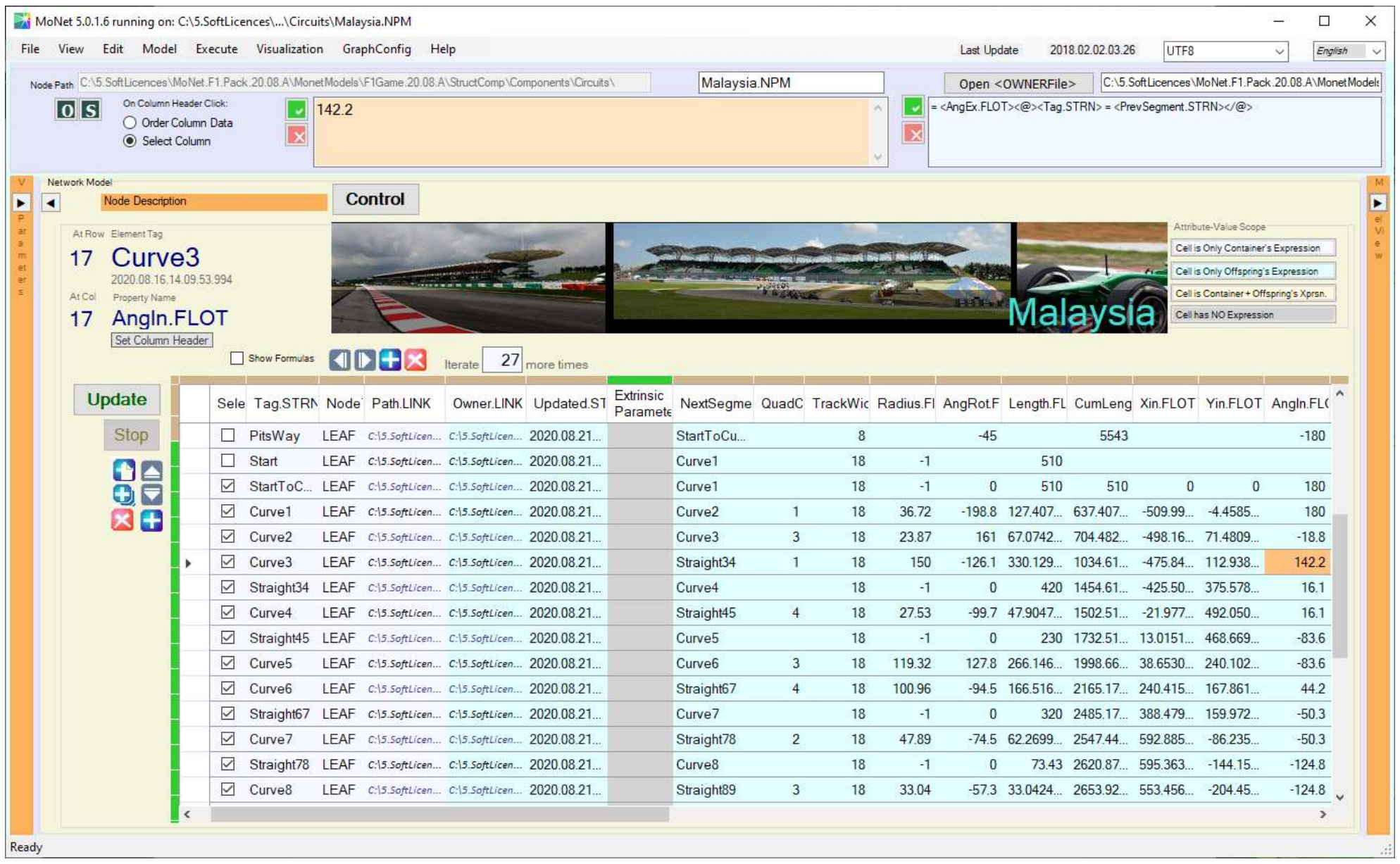

**Figure A1.** Form of MoNet system showing the simulation of the circuit of Sepang in Malaysia. This panel shows the description of the elements (straights and curves) of the circuit. The circuit component also receives information about the racing car and the race plan needed to simulate each stint (segment) of a race.

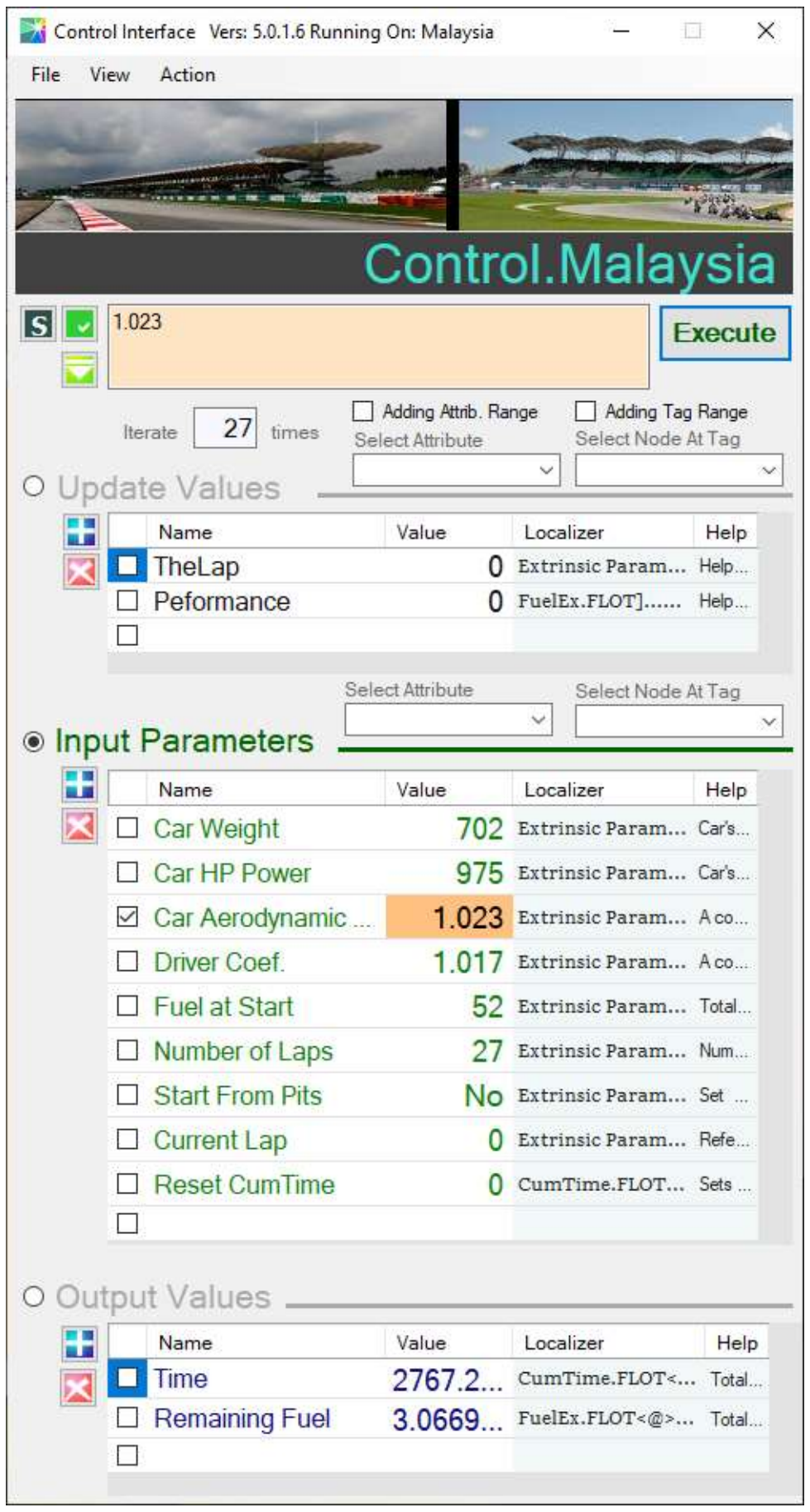

**Figure A2. The** Form that allows the MoNet's user to control the performance of a car-race plan-circuit conjunction shown in Figure A1.   .

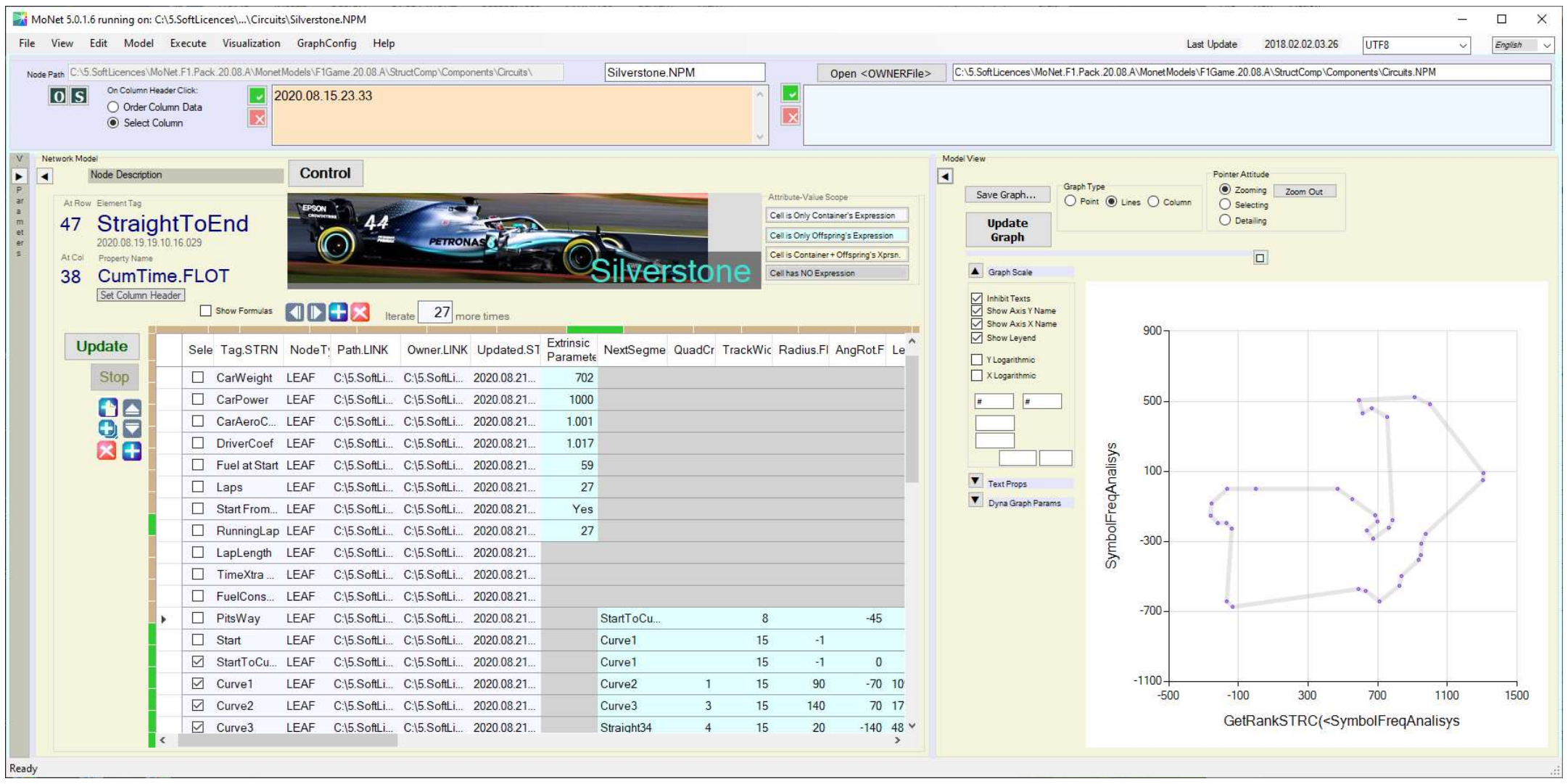

**Figure A3.** Form of MoNet system showing the simulation of the circuit of Silverstone in England. This panel shows the description of the elements (straights and curves) of the circuit. In this representation a panel to show the circuit shape is open.

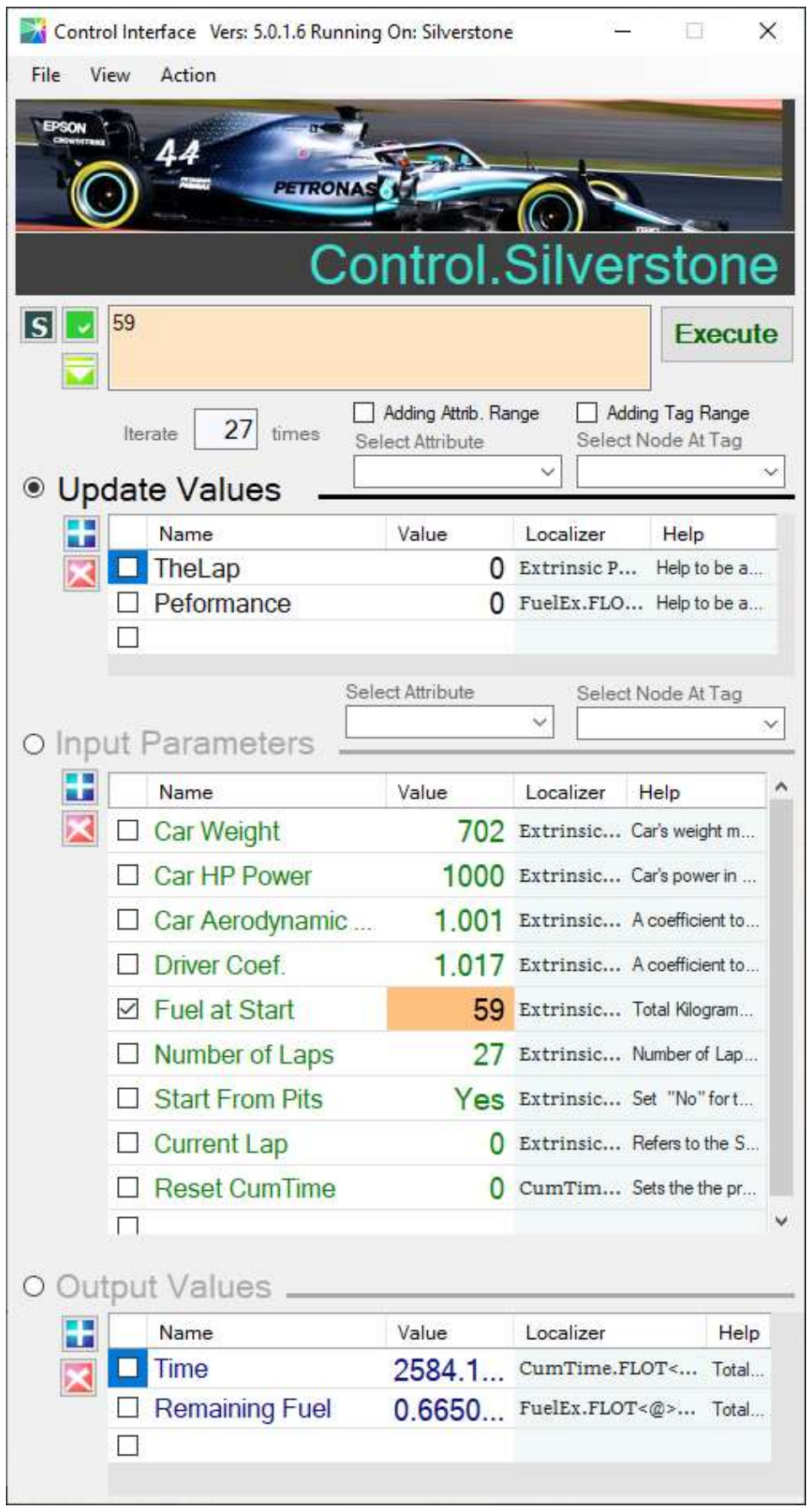

**Figure A3.** The Form that allows the MoNet's user to control the performance of a car-race plan-circuit conjunction shown in Figure A1. .

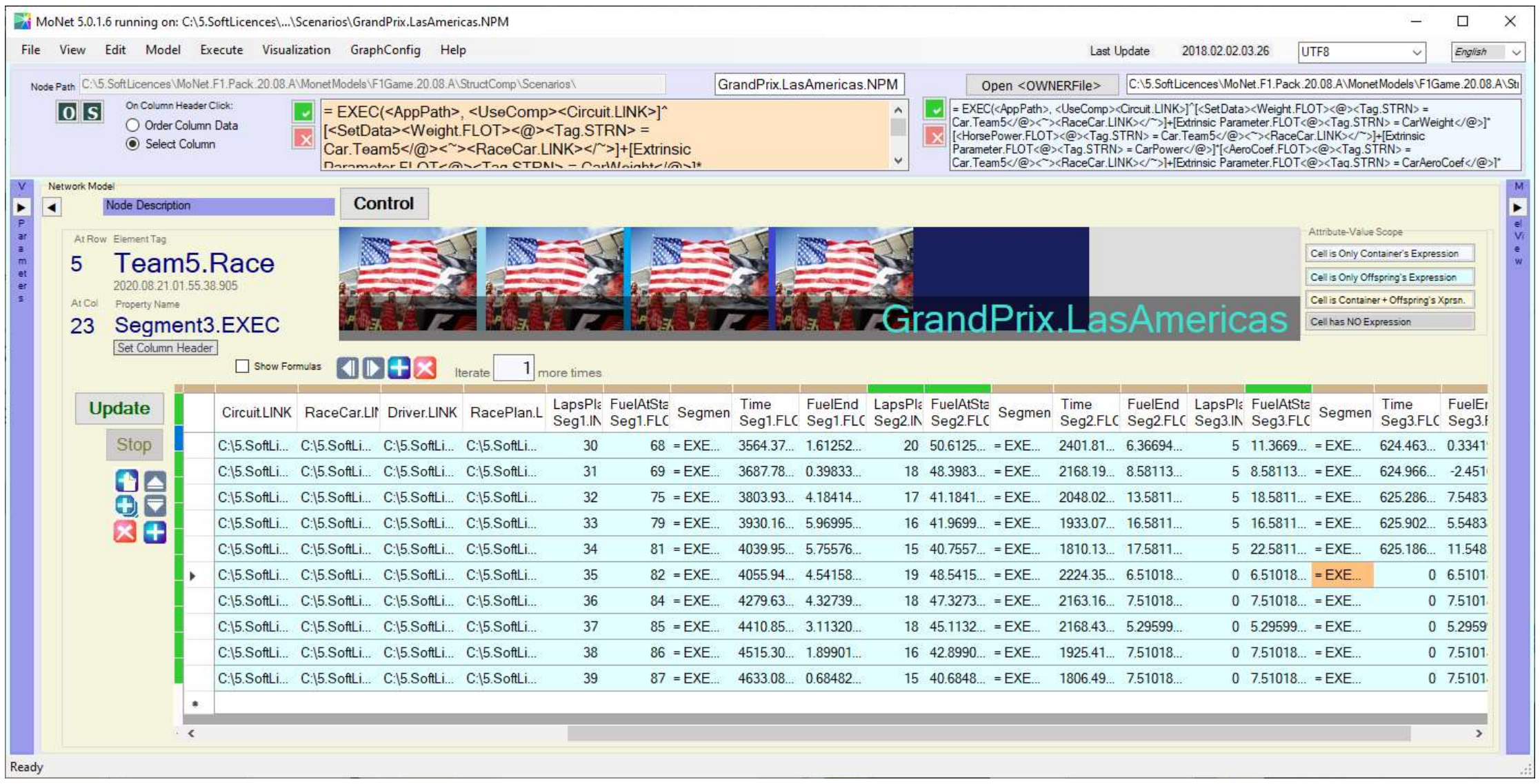

**Figure A4.** The section of the system collecting the results of the Teams for the race segments in the Grand Prix of USA in the circuit of Las Americas.